\documentclass[sts,preprint]{imsart-custom}

\usepackage{times}
\usepackage{graphicx}
\usepackage{subfigure} 

\usepackage{natbib}

\usepackage{hyperref}

\usepackage{amsmath}
\usepackage{amssymb}
\usepackage{amsthm}
\newcommand{\Xe}{\vec{X}_\mathrm{e}  }
\newcommand{\XeT}{\vec{X}_\mathrm{e}^\top}

\newcommand{\ye}{\begin{bmatrix} \vec{y}  \\ \vec{y}_\mathrm{u} \end{bmatrix}}
\newcommand{\G}{\left(\Xe^\top \Xe \right)^{-1}}

\newtheorem{theorem}{Theorem}
\renewcommand{\vec}[1]{\mathbf{#1}}
\DeclareMathOperator*{\argmin}{arg\,min}

\begin{document}

\begin{frontmatter}
\title{Projected Estimators for Robust Semi-supervised Classification}
\runtitle{Projected Estimators for Robust Semi-supervised Classification}

\begin{aug}
  \author{Jesse H. Krijthe\ead[label=e1]{jkrijthe@gmail.com}}
  \and
  \author{Marco Loog\ead[label=e2]{m.loog@tudelft.nl}}

  \address{Pattern Recognition \& Bioinformatics, Delft University of Technology \& Department of Molecular Epidemiology, Leiden University Medical Center\printead{e1}}

  \address{Pattern Recognition \& Bioinformatics, Delft University of Technology \& Image Group, University of Copenhagen \printead{e2}}
  \runauthor{J.H. Krijthe \& M. Loog}
\end{aug}

\begin{abstract}
For semi-supervised techniques to be applied safely in practice we at least want methods to outperform their supervised counterparts. We study this question for classification using the well-known quadratic surrogate loss function. Using a projection of the supervised estimate onto a set of constraints imposed by the unlabeled data, we find we can safely improve over the supervised solution in terms of this quadratic loss. Unlike other approaches to semi-supervised learning, the procedure does not rely on assumptions that are not intrinsic to the classifier at hand. It is theoretically demonstrated that, measured on the labeled and unlabeled training data, this semi-supervised procedure never gives a lower quadratic loss than the supervised alternative.  To our knowledge this is the first approach that offers such strong, albeit conservative, guarantees for improvement over the supervised solution. The characteristics of our approach are explicated using benchmark datasets to further understand the similarities and differences between the quadratic loss criterion used in the theoretical results and the classification accuracy often considered in practice.
\end{abstract}

\end{frontmatter}

\section{Introduction}
\label{Introduction}
We consider the problem of semi-supervised classification using the quadratic loss function, which is also known as least squares classification or Fisher's linear discriminant classification \citep{Hastie2009,Poggio2003}. Suppose we are given an $N_l \times d$ matrix with feature vectors $\vec{X}$, labels $\vec{y} \in \{0,1\}^{N_l}$ and an $N_u \times d$  matrix with unlabeled objects $\vec{X_u}$ from the same distribution as the labeled objects. The goal of semi-supervised learning is to improve the classification decision function $f: \mathbb{R}^d \to \mathbb{R}$ using the unlabeled information in $\vec{X_u}$ as compared to the case where we do not have these unlabeled objects. In this work, we focus on linear classifiers where $f(\vec{x})=\vec{w}^T \vec{x}$. 

Much work has been done on semi-supervised classification, in particular on what additional assumptions about the unlabeled data may help improve classification performance. These additional assumptions, while successful in some settings, are less successful in others where they do not hold. In effect they can greatly deteriorate performance when compared to a supervised alternative \citep{Cozman2006}. Since, in semi-supervised applications, the number of labeled objects may be small, the effect of these assumptions is often untestable. In this work, we introduce a conservative approach to training a semi-supervised version of the least squares classifier that is guaranteed to improve over the supervised least squares classifier, in terms of the quadratic loss on the labeled and unlabeled examples. It is the first procedure for which it is possible to give strong guarantees of non-degradation of this type (Theorem~\ref{th:robustness}).

To guarantee these improvements, we avoid additional assumptions altogether. We introduce a constraint set of parameter vectors induced by the unlabeled data, which does not rely on additional assumptions about the data. Using a projection of the supervised solution vector onto this constraint set, we derive a method that can be proven to never degrade the surrogate loss evaluated on the labeled and unlabeled training data when compared to the supervised solution. Experimental results indicate that it not only never degrades, but often improves performance. Our experiments also indicate the results hold when performance is evaluated on objects in a test set that were not used as unlabeled objects during training.

Others have attempted to mitigate the problem of reduction in performance in semi-supervised learning by introducing safe versions of semi-supervised learners \citep{Li2011,Loog2010,Loog2014a}. These procedures do not offer any guarantees or only do so once particular assumptions about the data hold. Moreover, the proposed method can be formulated as a convex quadratic programming problem which can be solved using a simple gradient descent procedure.


The rest of this work is organized as follows. The next section discusses related work. Section \ref{section:projections} introduces our projection approach to semi-supervised learning. Section \ref{section:theory} discusses the theoretical performance guarantee and its implications. Section \ref{section:interpretations} provides some alternative interpretations of the method and relations to other approaches. In Section \ref{section:empirical} empirical illustrations on benchmark datasets are presented to understand how the theoretical results in terms of quadratic loss in Section \ref{section:theory} relate to classification error. We end with a discussion of the results and conclude.

\section{Prior Work and Assumptions}

Early work on semi-supervised learning dealt with the missing labels through the use of Expectation Maximization in generative models or closely related self-learning \citep{McLachlan1975}. Self-learning is a simple wrapper method around any supervised procedure. Starting with a supervised learner trained only on the labeled objects, we predict labels for the unlabeled objects. Using the known labels and the predicted labels for the unlabeled objects, or potentially the predicted labels with highest confidence, we retrain the supervised learner. This process is iterated until the predicted labels converge. Although simple, this procedure has seen some practical success \citep{Nigam2000}.

More recent work on semi-supervised methods involves either the assumption that the decision boundary is in a low-density region of the feature space, or that the data is concentrated on a low-dimensional manifold. A well-known procedure using the first assumption is the Transductive SVM \citep{Joachims1999}. It can be interpreted as minimizing the following objective:
\begin{multline}
\label{eq:TSVM}
\min_{\vec{w} \in \mathbb{R}^d,\vec{y}_\text{u} \in \{-1,+1\}^{N_u}} \sum_{i=1}^{N_l} \max(1-\vec{y}_i \vec{w}^\top \vec{x},0) + \lambda ||\vec{w}||^2 \\ + \lambda_u \sum_{i=1}^{N_u} \max(1-\vec{y}_\text{u}^{(i)} \vec{w}^\top \vec{x},0)
\end{multline}
where class labels are encoded using $+1$ and $-1$. This leads to a hard to optimize, non-convex, problem, due to the dependence on the labels of the unlabeled objects $\vec{y}_\text{u}$. Others, such as \citep{Sindhwani2006}, have proposed procedures to efficiently find a good local minimum of this function. Similar low-density ideas have been proposed for other classifiers, such as entropy regularization for logistic regression \citep{Grandvalet2005} and a method for Gaussian processes \citep{Lawrence2004}. One challenge with these procedures is setting the additional parameter $\lambda_u$ that is introduced to control the effect of the unlabeled objects. This is both a computational problem, since minimizing \eqref{eq:TSVM} is already hard for a single choice of $\lambda_u$, as well as a estimation problem. If the parameter is incorrectly set using, for example, cross-validation on a limited set of labeled examples, the procedure may actually reduce performance as compared to a supervised SVM which disregards the unlabeled data. It is this behaviour that our procedure avoids. While it may be outperformed by the TSVM if the low-density assumption holds, robustness against deterioration would still constitute an important property in the cases when we are not sure whether it does hold.

An attempt at safety in semi-supervised learning was introduced in \citep{Li2011}, who propose a safe variant for semi-supervised support vector machines. By constructing a set of possible decision boundaries using the unlabeled and labeled data, the decision boundary is chosen that is least likely to degrade performance. The goal of our work is also similar to that of \citep{Loog2010,Loog2014a}, who introduce a semi-supervised version of linear discriminant analysis, which is closely related to the least squares classifier considered here. There, explicit constraints are proposed that take into account the unlabeled data. In our work, these constraints need not be explicitly derived, but follow  directly from the choice of loss function and the data. While the impetus for these works is similar to ours, they provide no theory to guarantee no degradation in performance will occur similar to our results in Section \ref{section:theory}.

\section{Projection Method}
\label{section:projections}
The proposed projection method works by forming a constraint set of parameter vectors $\Theta$, informed by the labeled \emph{and unlabeled} objects, that is guaranteed to include $\vec{w}_\text{oracle}$, the solution we would obtain if we had labels for all the training data. We will then find the closest projection of the supervised solution $\vec{w}_{\text{sup}}$ onto this set, using a chosen distance measure. This new estimate, $\vec{w}_{\text{semi}}$, will then be guaranteed to be closer to the oracle solution than the supervised solution $\vec{w}_{\text{sup}}$ in terms of this distance measure. For a particular choice of measure, it follows (Section \ref{section:theory}) that $\vec{w}_{\text{semi}}$ will always have lower quadratic loss when measured on the labeled and unlabeled training data, as compared to $\vec{w}_{\text{sup}}$.
Before we move to our particular contribution, we first introduce briefly the standard supervised least squares classifier.

\subsection{Supervised Solution}
We consider classification using a quadratic surrogate loss \citep{Hastie2009}. In the supervised setting, the following objective is minimized for $\vec{w}$:
\begin{equation}
\label{eq:supervisedloss}
L(\vec{w},\vec{X},\vec{y}) = \lVert \vec{X} \vec{w} - \vec{y} \rVert^2
\end{equation}
The supervised solution $\vec{w}_{\text{sup}}$ is given by the minimization of \eqref{eq:supervisedloss} for $\vec{w}$. The well-known closed form solution to this problem is given by
\begin{equation}
\label{eq:supervisedsolution}
\vec{w}_{\text{sup}} = (\vec{X}^\top \vec{X})^{-1} \vec{X}^\top \vec{y}
\end{equation}
If the true labels corresponding to the unlabeled objects, $\vec{y}_\text{u}^{\ast}$, would be given, we could incorporate these by extending the vector of labels ${\vec{y}_\text{e}^\ast}^\top = \left[ \vec{y}^\top {\vec{y}_\text{u}^\ast}^\top \right]$ as well as the design matrix $\vec{X}_\text{e}^\top = \left[ \vec{X}^\top \vec{X}_\text{u}^\top \right]$ and minimize $L(\vec{w},\Xe, \vec{y}_\text{e}^\ast)$ over the labeled as well as the unlabeled objects. We will refer to this oracle solution as $\vec{w}_\text{oracle}$. 

\subsection{Constraint Set}
Our proposed semi-supervised approach is to project the supervised solution $\vec{w}_\text{sup}$ onto the set of all possible classifiers we would be able to get from some labeling of the unlabeled data. To form the constraint set, consider all possible labels for the unlabeled objects $\vec{y}_\text{u} \in [0,1]^{N_u}$. This includes fractional labelings, where an objects is partly assigned to class $0$ and partly to class $1$. For instance, $0.5$ indicates the object is equally assigned to both classes. For a particular labeling $\vec{y}_\text{e}^\top = \left[ \vec{y}^\top \vec{y}_\text{u}^\top \right]$, we can find the corresponding parameter vector by minimizing $L(\vec{w},\vec{X}_\text{e},\vec{y}_\text{e})$ for $\vec{w}$.
This objective remains the same as \eqref{eq:supervisedloss} except that fractional labels are now also allowed. Minimizing the objective for all possible labelings generates the following set of solutions:
\begin{equation}
\label{eq:constrainedregion}
\Theta=\left\{ \G \XeT \ye \mid \vec{y}_\text{u} \in [0,1]^{N_u} \right\} \, .
\end{equation}
Note that this set, by construction, will also contain the solution $\vec{w}_\text{oracle}$, corresponding to the true but unknown labeling $\vec{y}_\text{e}^{\ast}$. Typically, $\vec{w}_\text{oracle}$ is a better solution than $\vec{w}_\text{sup}$ and so we would like to find a solution more similar to $\vec{w}_\text{oracle}$. This can be accomplished by projecting $\vec{w}_\text{sup}$ onto $\Theta$.

\subsection{Choice of Metric}
It remains to determine how to calculate the distance between $\vec{w}_\text{sup}$ and any other $\vec{w}$ in the space. We will consider the following metric:
\begin{equation}
\label{eq:metric}
\text{d}(\vec{w},\vec{w}^\prime)=\sqrt{\left( \vec{w}-\vec{w}^\prime \right)^\top \vec{X}_{\circ}^\top \vec{X}_{\circ}  \left( \vec{w}-\vec{w}^\prime \right)}
\end{equation}
where we assume $\vec{X}_{\circ}^\top \vec{X}_{\circ}$ is a positive definite matrix. The projected estimator can now be found by minimizing this distance between the supervised solution and solutions in the constraint set:
\begin{equation}
\label{eq:projection}
\vec{w}_\mathrm{semi} = \min_{\vec{w} \in \Theta} \text{d}(\vec{w},\vec{w}_\text{sup})
\end{equation}
Setting $\vec{X}_\circ=\Xe$ measures the distances using both the labeled and unlabeled data. This choice has the desirable theoretical properties leading us to the sought-after improvement guarantees as we will demonstrate in Section~\ref{section:theory}.

\subsection{Optimization}
By plugging into \eqref{eq:projection} the closed form solution of $\vec{w}_\text{sup}$ and $\vec{w}$  for a given $\vec{y}_\text{u}$, this problem can be written as a convex minimization problem in terms of $\vec{y}_\text{u}$, the unknown, fractional labels of the unlabeled data. This results in a quadratic programming problem, which can be solved using a simple gradient descent procedure that takes into account the constraint that the labels are within $[0,1]$. The solution of this quadratic programming problem $\vec{\hat{y}}_\text{u}$ can then be used to find  $\vec{w}_\text{semi}$ by treating these imputed labels as the true labels of the unlabeled objects and combining them with the labeled examples in Equation \eqref{eq:supervisedsolution}.

\section{Theoretical Analysis}
\label{section:theory}

We start by stating and proving our main result which is a guarantee of non-degradation in performance of the proposed method compared to the supervised classifier. We then discuss extensions of this result to other settings and give an indication of when improvement over the supervised solution can be expected.

\subsection{Robustness Guarantee}
\begin{theorem}
\label{th:robustness}
Given $\vec{X}$, $\vec{X}_\mathrm{u}$ and $\vec{y}$, $\Xe^\top \Xe$ positive definite and $\vec{w}_\mathrm{sup}$ given by \eqref{eq:supervisedsolution}. For the projected estimator $\vec{w}_\mathrm{semi}$ proposed in \eqref{eq:projection}, the following result holds:
$$L(\vec{w}_\mathrm{semi},\Xe,\vec{y}_\mathrm{e}^{\ast}) \leq L(\vec{w}_\mathrm{sup},\Xe,\vec{y}_\mathrm{e}^{\ast}) $$
\end{theorem}
In other words: $\vec{w}_\text{semi}$ will \emph{always} be at least as good or better than $\vec{w}_\text{sup}$, in terms of the quadratic surrogate loss on all, labeled and unlabeled, training data.
\begin{proof}
The proof of this result follows from a geometric interpretation of our procedure. Consider the following inner product that induces the distance metric in Equation \eqref{eq:metric}:
\begin{equation}
\left\langle \vec{w}, \vec{w}^\prime \right\rangle = \vec{w}^\top \vec{X}_\text{e}^\top \vec{X}_\text{e} \vec{w}^\prime \,. \nonumber
\end{equation}
Let $\mathcal{H}_{\vec{X}_\text{e}} = ( \mathbb{R}^d,\left\langle ., . \right\rangle )$ be the inner product space corresponding with this inner product. As long as $\XeT \Xe$ is positive definite, this is a Hilbert space. Next, note that the constraint space $\Theta$ is convex. More precisely, because, for any $k \in [0,1]$ and $\vec{w}_\text{1},\vec{w}_\text{2} \in \Theta$ we have that
\begin{align}
(1-k) \vec{w}_\text{1} + k \vec{w}_\text{2}  = & (1-k) \G \XeT \left[\vec{y}^\top \vec{y}_\text{1}^\top \right] \nonumber \\
 & + k \G \XeT \left[\vec{y}^\top \vec{y}_\text{2}^\top \right] \nonumber \\ 
 = & \G \XeT \left[\vec{y}^\top ~~ k \vec{y}_\text{1}^\top + (1-k) \vec{y}_\text{2}^\top \right] \nonumber \\
 \in & \; \Theta \nonumber
\end{align}

where the last statement holds because $k \vec{y}_\text{1}^\top + (1-k) \vec{y}_\text{2}^\top \in [0,1]^{N_u}$.

By construction $\vec{w}_\text{semi}$ is the closest projection of $\vec{w}_\text{sup}$ onto this convex constraint set $\Theta$ in $\mathcal{H}_{\vec{X}_\text{e}}$. One of the properties for projections onto a convex subspace in a Hilbert space is \citep[Proposition 1.4.1.]{Aubin2000} that 
\begin{equation}
\label{eq:projectiontheorem}
\text{d}(\vec{w}_\text{semi},\vec{w}) \leq \text{d}(\vec{w}_\text{sup},\vec{w})
\end{equation}
for any $\vec{w} \in \Theta$. In particular consider $\vec{w}=\vec{w}_\text{oracle}$, which by construction is within $\Theta$. That is, all possible labelings correspond to an element in $\Theta$, so this also holds for the true labeling $\vec{y}_\text{u}^\ast$. Plugging in the closed form solution of $\vec{w}_\text{oracle}$ into \eqref{eq:projectiontheorem} we find:
\begin{flalign}
\text{d}(\vec{w}_\text{semi},\vec{w}_\text{oracle})^2 = & \vec{w}_\text{semi}^\top \Xe^\top \Xe \vec{w}_\text{semi} \nonumber \\ \nonumber
& - 2 \vec{w}_\text{semi}^\top \Xe^\top {\vec{y}_\text{e}^\ast} + {\vec{y}_\text{e}^\ast}^\top {\vec{y}_\text{e}^\ast}\\ \nonumber
& +  C\\ \nonumber
= & L(\vec{w}_\text{semi},\Xe,\vec{y}_\text{e}^{\ast}) + C\\ \nonumber
\end{flalign}
and
\begin{flalign}
\text{d}(\vec{w}_\text{sup},\vec{w}_\text{oracle})^2 = & \vec{w}_\text{sup}^\top \Xe^\top \Xe \vec{w}_\text{sup} \nonumber \\ \nonumber
& - 2 \vec{w}_\text{sup}^\top \Xe^\top {\vec{y}_\text{e}^\ast} +  {\vec{y}_\text{e}^\ast}^\top {\vec{y}_\text{e}^\ast} \\ \nonumber
& +  C\\ \nonumber
= & L(\vec{w}_\text{sup},\Xe,\vec{y}_\text{e}^{\ast}) + C  \\ \nonumber
\end{flalign}
where $C$ is the same constant in both cases. From this the result in Theorem \ref{th:robustness} follows directly.
\end{proof}

\subsection{Transduction and Regularization}
It is possible to derive a similar result for performance improvement on the unlabeled data alone by using $\vec{X}_\circ=\vec{X}_\text{u}$ in the distance measure and changing the constrained hypothesis space to:
\begin{equation}
\label{eq:constrainedregion2}
\Theta_\text{u} = \left\{ (\vec{X}_\mathrm{u}^\top \vec{X}_\mathrm{u})^{-1} \vec{X}_\mathrm{u}^\top \vec{y}_\mathrm{u} \mid \vec{y}_\text{u} \in [0,1]^{N_u} \right\}  \, . \nonumber
\end{equation}
This would lead to a guarantee of the form:
\begin{equation}
L(\vec{w}_\mathrm{semi},\vec{X}_\mathrm{u},\vec{y}_\mathrm{u}^{\ast}) \leq L(\vec{w}_\mathrm{sup},\vec{X}_\mathrm{u},\vec{y}_\mathrm{u}^{\ast})  \, . \nonumber
\end{equation}
However, since we would not just like to perform well on the given unlabeled data, but on unseen data from the same distribution as well, we include the labeled data in the construction of the constrained hypothesis space.

The result in Theorem \ref{th:robustness} also holds if we include regularization in the supervised classifier. Using $L_2$ regularization, the supervised solution becomes:
\begin{equation}
\label{eq:regsupervisedsolution}
\vec{w}_{\text{sup}} = (\vec{X}^\top \vec{X} + \lambda \vec{I})^{-1} \vec{X}^\top \vec{y} \nonumber
\end{equation}
where $\lambda$ is a regularization parameter and $\vec{I}$ a $d \times d$ identity matrix, potentially containing a $0$ for the diagonal entry corresponding to the constant feature that encodes the bias. Theorem \ref{th:robustness} also holds for this regularized supervised estimator.

\subsection{Improved Performance}
Since the inequality in Theorem \ref{th:robustness} is not necessarily a strict inequality, it is important to get an idea when we can expect improvement of the semi-supervised learner, rather than just equality of the losses. Consider a single unlabeled object. Improvement happens whenever $\vec{w}_\text{sup} \neq \vec{w}_\text{semi}$, which occurs if $\vec{w}_\text{sup} \notin \Theta$.  For this to occur it needs to be impossible to assign labels $\text{y}_\text{u}$ such that we can retrieve the $\vec{w}_\text{sup}$ by minimizing $L(\vec{w},\vec{X}_\text{e},\vec{y}_\text{e})$. This in turn occurs when there is no $\text{y}_\text{u} \in [0,1]$ for which the gradient

\begin{equation}
\nabla \lVert \vec{X}_\text{e} \vec{w} - \vec{y}_\text{e} \rVert^2\bigg|_{\vec{w}=\vec{w}_\text{sup}}=\vec{0}  \, . \nonumber
\end{equation}

This happens only if $\vec{x}_\text{u}^\top \vec{w}_\text{sup} > 1$ or $\vec{x}_\text{u}^\top \vec{w}_\text{sup} < 0$. In other words, if observations $\vec{x}_\text{u}$ are possible with values that are sufficiently large (or small) and $\vec{w}_\text{sup}$ is not small enough to mitigate this, an update will occur. For many datasets, we might expect this to be true for at least one observation in a large set of unlabeled objects. This is especially true if the  supervised solution is not sufficiently regularized and the $\vec{x}_\text{u}^\top \vec{w}_\text{sup}$ can easily be larger than $1$ or smaller than $0$. The experiments in Section \ref{section:empirical} indeed confirm that generally improvements can be expected by means of the proposed semi-supervised learning strategy.

\section{Relation to Other Methods}
\label{section:interpretations}
The projection method in Equation \eqref{eq:projection}, using $\vec{X}_{\circ}=\Xe$ in the distance measure, can be rewritten in a different form:
\begin{equation}
\argmin_{\vec{w}_\text{semi}} \max_{\vec{y}_\text{u} \in [0,1]^{N_u}} L(\vec{w}_\text{semi},\Xe,\vec{y}_\text{e}) - L(\vec{w}_\text{sup},\Xe,\vec{y}_\text{e}) \nonumber
\end{equation}
In other words, the procedure can be interpreted as a minimization of the difference in loss on the labeled and unlabeled data between the new solution and the supervised solution, over all possible labelings of the unlabeled data. From this perspective the projected estimator is similar to Maximum Contrastive Pessimistic Likelihood Estimation proposed by \citet{Loog2016} who consider using log likelihood as the loss function. In this formulation it is apparent that the projected estimator is very conservative, since it has to have low loss for all possible labelings, even very unlikely ones.

In a similar way an alternative choice of distance function, $\vec{X}_{\circ}=\vec{X}$, has a different interpretation. It is the minimizer of the supervised loss function under the constraint that its solution has to be a minimizer for some labeling of the unlabeled data:
\begin{equation}
\argmin_{\vec{w} \in \Theta} L(\vec{w},\vec{X},\vec{y}) \nonumber
\end{equation}
with $\Theta$ defined as in Equation \eqref{eq:constrainedregion}. This formulation corresponds to the Implicitly Constrained Least Squares Classifier \citep{Krijthe2015} and seems less conservative since the solution does not need to have a low loss for all possible labelings, it merely has to work well on the labeled examples. For this distance measure, the proof in Section \ref{section:theory} no longer holds, but empirical results indicate it may have better performance in practice, while it still protects against deterioration in performance by minimizing the loss over only the labeled objects.

Another interpretation of the projection procedure is that it minimizes the squared difference between the predictions of the supervised solution and a new semi-supervised solution on the set of objects in $\vec{X}_{\circ}$, while ensuring the semi-supervised solution corresponds to a possible labeling of the unlabeled objects:
\begin{equation}
\min_{\vec{w} \in \Theta} \lVert \vec{X}_{\circ} \vec{w} - \vec{X}_{\circ} \vec{w}_\text{sup} \lVert^2 \,.\nonumber
\end{equation}
Since this comparison requires only the features in $\vec{X}_{\circ}$ and not the corresponding labels, this can be done either on the labeled data, when we choose $\vec{X}_{\circ}=\vec{X}$, but also on the labeled and unlabeled data combined when $\vec{X}_{\circ}=\Xe$. This interpretation is similar to the work of \cite{Schuurmans2002}, where the unlabeled objects are also used to measure the difference in predictions of two hypotheses. 

\section{Experimental Analysis}
\label{section:empirical}
For our experiments, we consider $16$ classification datasets. $6$ of these are the semi-supervised learning benchmark datasets proposed by \citet{Chapelle2006}, while the other $10$ were retrieved from the UCI Machine Learning repository \citep{Bache2013}. All of the datasets are binary classification problems, or were turned into two-class problems by merging several similar classes. As a preprocessing step, missing input values were imputed using medians and modes for the Mammography and Diabetes datasets. The code to reproduce the results presented here is available from the first author's website.

The number of labeled examples is chosen such that $N_l>d$. This is necessarily to have a high probability that the matrix $\vec{X}_\text{e}^\top \vec{X}_\text{e}$ is positive definite, which was a requirement of Theorem \ref{th:robustness}. More importantly, this avoids peaking behaviour \citep{Raudys1998, Opper1996}, were the unregularized supervised least squares classifier has low performance when the matrix $\vec{X}^\top \vec{X}$ is not full-rank. For the SVM and TSVM implementations we made use of the SVMlin software \citep{Sindhwani2006}. For these we used parameter settings $\lambda=0.01$ and $\lambda_u=1$.

\subsection{Robustness}
\begin{figure*}
\centering
\includegraphics[scale=0.45]{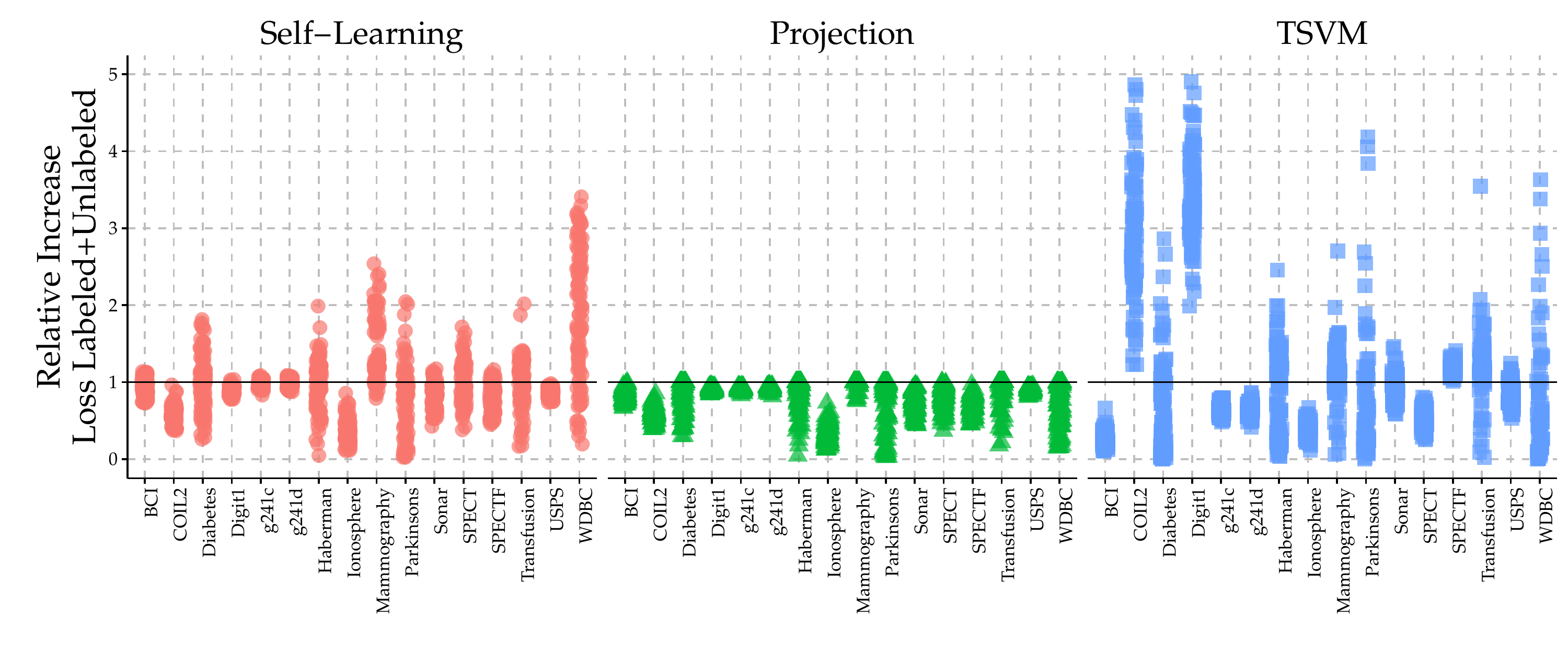}
\caption{Ratio of the loss in terms of surrogate loss of supervised and semi-supervised solutions measured on the labeled and unlabeled instances. Values smaller than $1$ indicate that the semi-supervised method gives a lower average surrogate loss than its supervised counterpart. For both the projected estimator and self-learning this supervised counterpart is the supervised least squares classifier and loss is in terms of quadratic loss. For the $L_2$-Transductive SVM, quadratic hinge loss is used and compared to the quadratic hinge loss of a supervised $L_2$-SVM. Unlike the other semi-supervised procedures, the projection method, evaluated on labeled and unlabeled data, never has higher loss than the supervised procedure, as was proven in Theorem~\ref{th:robustness}.}
\label{fig:lossdifference}
\end{figure*}

To illustrate Theorem \ref{th:robustness} experimentally, as well as study the performance of the proposed procedure on a test set, we set up the following experiment. For each of the $16$ datasets, we randomly select $2 d$ labeled objects. We then randomly sample, with replacement, $1000$ objects as the unlabeled objects from the dataset. In addition, a test set of $1000$ objects is also sampled with replacement. This procedure is repeated $100$ times and the ratio between the average quadratic losses for the supervised and the semi-supervised procedure $\tfrac{L(\vec{w}_\text{semi},\Xe,\vec{y}_\text{e}^{\ast})}{L(\vec{w}_\text{sup},\Xe,\vec{y}_\text{e}^{\ast})}$ is calculated. As stated by Theorem \ref{th:robustness}, this quantity should be smaller than $1$ for the Projection procedure. We do the same for self-learning applied to the least squares classifier and to an $L_2$-Transductive SVM, which we compare to the supervised $L_2$-SVM. The results are shown in Figure \ref{fig:lossdifference}. 

On the labeled and unlabeled data the loss of the projection method is lower than that of the supervised classifier in all of the resamplings taken from the original dataset. Compare this to the behaviour of the self-learner. While on average, the performance is quite similar on these datasets, on a particular sample from a dataset, self-learning may lead to a higher quadratic loss than the supervised solution. It is favourable to have no deterioration in every resampling because in practice one does not deal with resamplings from an empirical distribution, but rather with a single dataset. A semi-supervised procedure should ideally work on this particular dataset, rather than in expectation over all datasets that one might have observed. We see similar behaviour as self-learning for the difference in squared hinge loss between the $L_2$-SVM and the $L_2$-TSVM. While better parameter choices may improve the number of resamplings with improvements, this experiment illustrates that while semi-supervised methods may improve performance on average, for a particular sample from a dataset there is no guarantee like Theorem \ref{th:robustness} for the projected estimator. When looking at the difference in loss on an unseen test set, we find a similar results (not shown).

\subsection{Learning Curves}
\begin{figure*}
\centering
\includegraphics[scale=0.4]{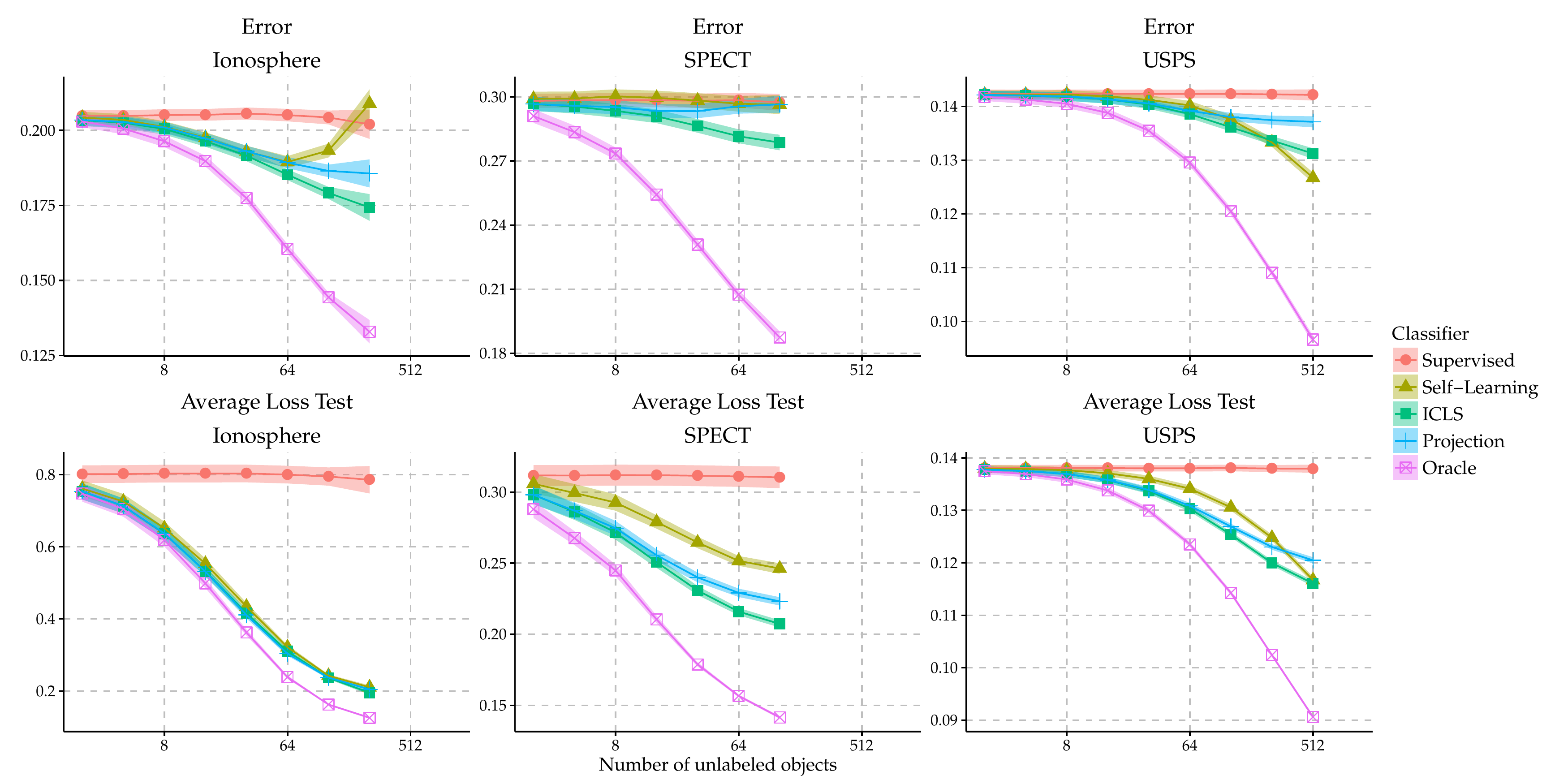}
\caption{Learning curves in terms of classification errors (top) and quadratic loss (bottom) on the test set for increasing numbers of \emph{unlabeled} data on three illustrative datasets. The lines indicate average errors respectively losses on the test set, averaged over $1000$ repeats. The shaded bars indicate $\pm 2$ standard errors around the mean.}
\label{fig:learningcurves}
\end{figure*}

To illustrate the behaviour of the procedure with increasing amounts of unlabeled data and to explore the relationship between the quadratic surrogate loss and classification accuracy we generate learning curves in the following manner. For each of three illustrative datasets (Ionosphere, SPECT and USPS), we randomly sample $2 d$ objects as labeled objects. The remaining objects are used as a test set. For increasing subsets of the unlabeled data $2,4,8,\dots,512$, randomly sampled without replacement, we train the supervised and semi-supervised learners and evaluate their performance on the test objects, in terms of classification accuracy as well as in terms of quadratic loss. We consider both the projection procedure where the distance measure is based on the labeled and the unlabeled data (denoted as Projection) as well as the projected estimator that only uses the labeled data in the distance measure (denoted as ICLS). The resampling is repeated $1000$ times and averages and standard errors are reported in Figure \ref{fig:learningcurves}.

The first dataset (Ionosphere) in Figure \ref{fig:learningcurves} is an example where the error of the self-learning procedure starts to increase once we add larger amounts of unlabeled data. In terms of the loss, however, the performance continues to increase. This illustrates that a decrease in the surrogate loss does not necessarily translates into a lower classification error. The projected estimators do not suffer from decrease in performance for larger numbers of unlabeled data in this example. In terms of the loss, however, there seems to be little difference between the three methods.

The second dataset (SPECT) is an example where both the self-learning procedure and the conservative projected estimator are not able to get any improvement out of the data, while the less conservative projection (ICLS) does show some improvement in terms of classification error.

On the USPS dataset the self-learning assumptions do seem to hold and it is able to attain a larger performance improvement as the amount of unlabeled data grows. Both in terms of the error and in terms of the loss, the projected estimators show smaller, but significant improvements.

\subsection{Cross-validation}
In a third experiment, we apply a cross-validation procedure to compare the performance increase in terms of the classification error of semi-supervised classifiers when compared to their supervised counterpart. The cross-validation experiments were set up as follows. For each dataset, the objects were split into 10-folds. Subsequently leaving out each fold, we combine the other 9 folds and randomly select $d+5$ labeled objects while the rest is used as unlabeled objects. We end up with a single prediction for each object, for which we evaluate the misclassification error. This procedure is repeated $20$ times and the averages are reported in Table 1.

The results indicate that in terms of classification errors, the projection procedure never significantly reduces performance over the supervised solution. This is in contrast to the self-learner, which does significantly increase classification error on $2$ of the datasets. The price the projected estimator pays for this robustness, is smaller improvements over the supervised classifier than the less conservative self-learner. The Transductive SVM shows similar behaviour as the self-learner: it shows large improvements over the supervised alternative, but is also prone to degradation in performance on other datasets. The ICLS procedure is, as expected, less conservative than the projection method based on the labeled and unlabeled observations, which leads to larger improvements on all of the datasets.

\begin{table*}[t]
\center
\caption{10-fold 20 repeat Cross-validation results for 16 datasets for the supervised least squares classifier, the projected least squares classifier (Projected), the projection based on only the labeled data (ICLS) and the self-learned least squares classifier. \textbf{Bold} respectively \underline{Underlined} values indicate whether the performance of a semi-supervised solution is significantly better or worse than the supervised alternative as evaluated by a one-sided Wilcoxon signed rank test with family wise error rate of $0.05$. The Win/Draw/Loss indicates on how many datasets a semi-supervised learner performs significantly better, equal or worse than the supervised alternative.}
\smallskip
\smallskip
\smallskip
\begin{tabular}{|l|llll|ll|}
   \hline
Dataset & Supervised & Self-Learning & ICLS & Projection & SVM & TSVM \\ 
  \hline
BCI & 0.40 $\pm$ 0.03 & \textbf{0.35 $\pm$ 0.02} & \textbf{0.28 $\pm$ 0.02} & \textbf{0.36 $\pm$ 0.03} & 0.30 $\pm$ 0.02 & 0.31 $\pm$ 0.02 \\ 
  COIL2 & 0.39 $\pm$ 0.01 & \textbf{0.26 $\pm$ 0.01} & \textbf{0.19 $\pm$ 0.01} & \textbf{0.34 $\pm$ 0.01} & 0.14 $\pm$ 0.01 & \underline{0.15 $\pm$ 0.01} \\ 
  Diabetes & 0.31 $\pm$ 0.02 & \underline{0.34 $\pm$ 0.01} & \textbf{0.30 $\pm$ 0.02} & \textbf{0.31 $\pm$ 0.02} & 0.36 $\pm$ 0.02 & \underline{0.38 $\pm$ 0.02} \\ 
  Digit1 & 0.42 $\pm$ 0.02 & \textbf{0.35 $\pm$ 0.02} & \textbf{0.20 $\pm$ 0.01} & \textbf{0.38 $\pm$ 0.01} & 0.06 $\pm$ 0.00 & 0.06 $\pm$ 0.01 \\ 
  g241c & 0.46 $\pm$ 0.01 & \textbf{0.39 $\pm$ 0.01} & \textbf{0.28 $\pm$ 0.01} & \textbf{0.42 $\pm$ 0.02} & 0.22 $\pm$ 0.01 & \textbf{0.21 $\pm$ 0.01} \\ 
  g241d & 0.44 $\pm$ 0.02 & \textbf{0.38 $\pm$ 0.01} & \textbf{0.29 $\pm$ 0.01} & \textbf{0.41 $\pm$ 0.02} & 0.23 $\pm$ 0.01 & \textbf{0.22 $\pm$ 0.01} \\ 
  Haberman & 0.29 $\pm$ 0.02 & 0.28 $\pm$ 0.02 & 0.29 $\pm$ 0.02 & 0.29 $\pm$ 0.02 & 0.29 $\pm$ 0.02 & 0.31 $\pm$ 0.03 \\ 
  Ionosphere & 0.28 $\pm$ 0.03 & \textbf{0.24 $\pm$ 0.01} & \textbf{0.19 $\pm$ 0.02} & \textbf{0.22 $\pm$ 0.03} & 0.20 $\pm$ 0.02 & 0.19 $\pm$ 0.02 \\ 
  Mammography & 0.30 $\pm$ 0.03 & 0.30 $\pm$ 0.02 & \textbf{0.29 $\pm$ 0.03} & 0.30 $\pm$ 0.03 & 0.30 $\pm$ 0.03 & 0.28 $\pm$ 0.02 \\ 
  Parkinsons & 0.25 $\pm$ 0.02 & 0.23 $\pm$ 0.03 & 0.24 $\pm$ 0.03 & 0.25 $\pm$ 0.03 & 0.22 $\pm$ 0.02 & 0.23 $\pm$ 0.02 \\ 
  Sonar & 0.44 $\pm$ 0.04 & \textbf{0.38 $\pm$ 0.04} & \textbf{0.33 $\pm$ 0.02} & \textbf{0.39 $\pm$ 0.02} & 0.26 $\pm$ 0.02 & \underline{0.33 $\pm$ 0.03} \\ 
  SPECT & 0.39 $\pm$ 0.04 & 0.38 $\pm$ 0.02 & \textbf{0.33 $\pm$ 0.03} & 0.39 $\pm$ 0.03 & 0.25 $\pm$ 0.03 & \textbf{0.20 $\pm$ 0.02} \\ 
  SPECTF & 0.44 $\pm$ 0.03 & \textbf{0.40 $\pm$ 0.04} & \textbf{0.36 $\pm$ 0.03} & \textbf{0.42 $\pm$ 0.03} & 0.25 $\pm$ 0.02 & \textbf{0.21 $\pm$ 0.01} \\ 
  Transfusion & 0.26 $\pm$ 0.02 & 0.28 $\pm$ 0.03 & 0.26 $\pm$ 0.02 & 0.26 $\pm$ 0.02 & 0.27 $\pm$ 0.01 & 0.28 $\pm$ 0.02 \\ 
  USPS & 0.42 $\pm$ 0.02 & \textbf{0.34 $\pm$ 0.02} & \textbf{0.20 $\pm$ 0.01} & \textbf{0.38 $\pm$ 0.02} & 0.11 $\pm$ 0.01 & \textbf{0.10 $\pm$ 0.00} \\ 
  WDBC & 0.09 $\pm$ 0.01 & \underline{0.13 $\pm$ 0.03} & \textbf{0.08 $\pm$ 0.01} & 0.09 $\pm$ 0.01 & 0.10 $\pm$ 0.01 & 0.11 $\pm$ 0.02 \\ 
   \hline
Total &  & 9 / 5 / 2 & 13 / 3 / 0 & 10 / 6 / 0 &  & 5 / 8 / 3 \\ 
   \hline
\end{tabular}

\end{table*}

\section{Discussion}
The main result of this work is summarized in Theorem \ref{th:robustness} and illustrated in Figure~\ref{fig:lossdifference}: the proposed semi-supervised classifier is guaranteed to improve over the supervised classifier in terms of the quadratic loss on all training data, labeled and unlabeled. The results from the experiments indicate that on average, both the projected estimator and other semi-supervised approaches often show improved performance, while on individual samples from the datasets, the projected estimator never reduces performance in terms of the surrogate loss. This is an important property since, in practical settings, one only has a single sample (i.e. dataset) from a classification problem, and it is important to know that performance will not be degraded when applying a semi-supervised version of a supervised procedure on that particular dataset. Even if we do not have enough labeled objects to accurately estimate this performance, Theorem \ref{th:robustness} guarantees we will not perform worse than the supervised alternative on the labeled and unlabeled data in terms of the surrogate loss. 

\subsection{Surrogate Loss}
Theorem~\ref{th:robustness} is limited to showing improvement in terms of quadratic loss. As the experiments also indicate, good properties in terms of this loss do not necessarily translate into good properties in terms of the error rate. In the empirical risk minimization framework, however, classifiers are constructed by minimizing surrogate losses. This particular semi-supervised learner is effective in terms of this objective. In this sense, it can be considered a proper semi-supervised version of the supervised quadratic loss minimizer.

One could question whether the quadratic loss is a good choice as surrogate loss \citep{Ben-David2012}. In practice, however, it can perform very well and is often on par and sometimes better than, for instance, an SVM employing hinge loss \citep{Rasmussen2005,Hastie2009,Poggio2003}. Moreover, the main result in this work basically demonstrates that strong improvement guarantees are at all possible for \emph{some} surrogate loss function. Whether and when an increase in performance in terms of this surrogate loss translates into improved classification accuracy is, like in the supervised setting, unclear. Much work is currently being done to understand the relationship between surrogate losses and 0-1 loss \citep{Bartlett2006, Ben-David2012}. 

\subsection{Conservatism}
Arguably, a robust semi-supervised learning procedure could also be arrived at by very conservatively setting the parameters controlling the influence of unlabeled data in semi-supervised learner procedures such as the TSVM. There are two reasons why this is difficult to achieve in practice. The first reason is a computational one. Most semi-supervised procedures are computationally intensive. Doing a grid search over both a regularization parameter as well as the parameter controlling the influence of the unlabeled objects using cross-validation is time-consuming. Secondly, and perhaps more importantly, it may be very difficult to choose a good parameter using limited labeled data. \citet{Goldberg2009} study this problem in more detail. While their conclusion suggests otherwise, their results indicate that performance degradation occurs on a significant number of datasets. 

The projected estimator presented here tries to alleviate these problems in two ways. Firstly, unlike many semi-supervised procedures, it can be formulated as a quadratic programming problem in terms of the unlabeled objects which has a global optimum (which is unique in terms of $\vec{w}$) and there are no hyper-parameters involved. Secondly, at least in terms of its surrogate loss, there is a guarantee performance will not be worse than the alternative of discarding the unlabeled data. 

As our results indicate, however, the proposed procedure is very conservative. The projection with $\mathbf{X}_\circ = \mathbf{X}$ (ICLS) is a classifier which is less conservative than the projection based on all data, and offers larger improvement in the experiments while still being robust to degradation of performance. For this procedure Theorem~\ref{th:robustness} does not hold.  Better understanding in what way we can still prove other robustness properties for this classifier is an open issue.

An alternative way to derive less conservative approaches could be by changing the constraint set $\Theta$. The purpose of this work has been to show that if we choose $\Theta$ conservatively, such that we can guarantee it contains the oracle solution $\vec{w}_\text{oracle}$, we can guarantee non-degradation, while still allowing for improved performance over the supervised solution in many cases. To construct a method with wider applicability, an interesting question is how to restrict $\Theta$ based on additional assumptions, while ensuring that $\vec{w}_\text{oracle} \in \Theta$ with high probability.



\section{Conclusion}
We introduced and analyzed an approach to semi-supervised learning with quadratic surrogate loss that has the interesting theoretical property of never decreasing performance when measured on the full, labeled and unlabeled, training set in terms of this surrogate loss when compared to the supervised classifier. This is achieved by projecting the solution vector of the supervised least squares classifier onto a constraint set of solutions defined by the unlabeled data. As we have illustrated through simulation experiments, the safe improvements in terms of the surrogate loss also partially translates into safe improvements in terms of the classification errors. Moreover, the procedure can be formulated as a standard quadratic programming problem, leading to a simple optimization procedure. An open problem is how to apply this procedure or a procedure with similar theoretical performance guarantees, to other loss functions.

\section*{Acknowledgements}
This work was funded by project P23 of the Dutch public-private research community COMMIT.

\bibliography{library}
\bibliographystyle{icml2016}

\end{document}